\documentclass[pdflatex,sn-nature]{sn-jnl}

\usepackage{graphicx}%
\usepackage{multirow}%
\usepackage{amsmath,amssymb,amsfonts}%
\usepackage{amsthm}%
\usepackage{mathrsfs}%
\usepackage[title]{appendix}%
\usepackage{xcolor}%
\usepackage{textcomp}%
\usepackage{manyfoot}%
\usepackage{booktabs}%
\usepackage{algorithm}%
\usepackage{algorithmicx}%
\usepackage{algpseudocode}%
\usepackage{listings}%
\usepackage{makecell}%
\usepackage{booktabs}
\usepackage{tabularx}
\usepackage{array}

\theoremstyle{thmstyleone}%
\theoremstyle{thmstyletwo}%

\theoremstyle{thmstylethree}%

\begin{document}

\title[Challenges of Auditing: Variability in Outputs of Large Language Models for Health]{Challenges of Auditing: Variability in Outputs of Large Language Models for Health}


\author[1]{\fnm{Yuan} \sur{Pu}}\email{yuan.pu@duke.edu}

\author[1]{\fnm{Yewon} \sur{Chang}}\email{yewon.chang@duke.edu}

\author[1]{\fnm{Furong} \sur{Jia}}\email{flora.jia@duke.edu}

\author[1]{\fnm{Xunjian} \sur{Yin}}\email{xunjian.yin@duke.edu}

\author[2,3]{\fnm{Jessica} \sur{Ma}}\email{jessica.ma@duke.edu}

\author[4]{\fnm{Ayman} \sur{Ali}}\email{ayman.ali@duke.edu}

\author[1,5]{\fnm{Monica} \sur{Agrawal}}\email{monica.agrawal@duke.edu}

\affil[1]{\orgdiv{Department of Computer Science}, \orgname{Duke University}, \orgaddress{\street{308 Research Drive}, \city{Durham}, \postcode{27708}, \state{NC}}}

\affil[2]{\orgdiv{Department of Medicine}, \orgname{Duke University}, \orgaddress{\street{40 Duke Medicine Circle}, \city{Durham}, \postcode{27710}, \state{NC}}}

\affil[3]{\orgdiv{Geriatrics and Extended Care}, \orgname{Durham VA Health System}, \orgaddress{\street{508 Fulton Street}, \city{Durham}, \postcode{27705}, \state{NC}}}

\affil[4]{\orgdiv{Department of Surgery}, \orgname{Duke University}, \orgaddress{\street{2301 Erwin Road}, \city{Durham}, \postcode{27710}, \state{NC}}}

\affil[5]{\orgdiv{Department of Biostatistics and Bioinformatics}, \orgname{Duke University}, \orgaddress{\street{2424 Erwin Road}, \city{Durham}, \postcode{27710}, \state{NC}}}


\abstract{
People increasingly use frontier AI models for health advice, but via different access modes (e.g., ChatGPT, ChatGPT Health, APIs) with varying settings. Here, we find systematic differences across access modes. Because evaluations typically rely on APIs while consumers interact through chatbot interfaces, these discrepancies limit evaluation validity. Our findings underscore an urgent need for model providers to enable faithful replication of consumer experiences and settings for rigorous audits.
}

\maketitle

\section{Background and Motivation}\label{background}

People are increasingly consulting large language models (LLMs) for health guidance, often through consumer-facing chatbots with vendor-built interfaces, such as ChatGPT  \cite{holzwarth2026tortoise, montero2026kff}. Given the high stakes underlying medical advice, there have been numerous attempts to evaluate the validity of LLM-generated responses. Some studies submit questions directly through \textit{the chatbot interface} and manually collect responses by copying-pasting \cite{ayers2023comparing, pan2023assessment, musheyev2024readability, draelos2026large}, an approach that mirrors consumer use but is difficult to scale, especially given the pace of model updates. The majority of evaluations therefore query the corresponding \textit{application programming interface (API)}, which enables programmatic, high-throughput access to the same underlying model \cite{jin2024better, fernandez2025evaluating, abrar2025empirical, sharma2025longitudinal, kopka2026evaluating, arora2025healthbench}.

\begin{figure}[ht]
    \centering
    \includegraphics[width=0.95\textwidth]{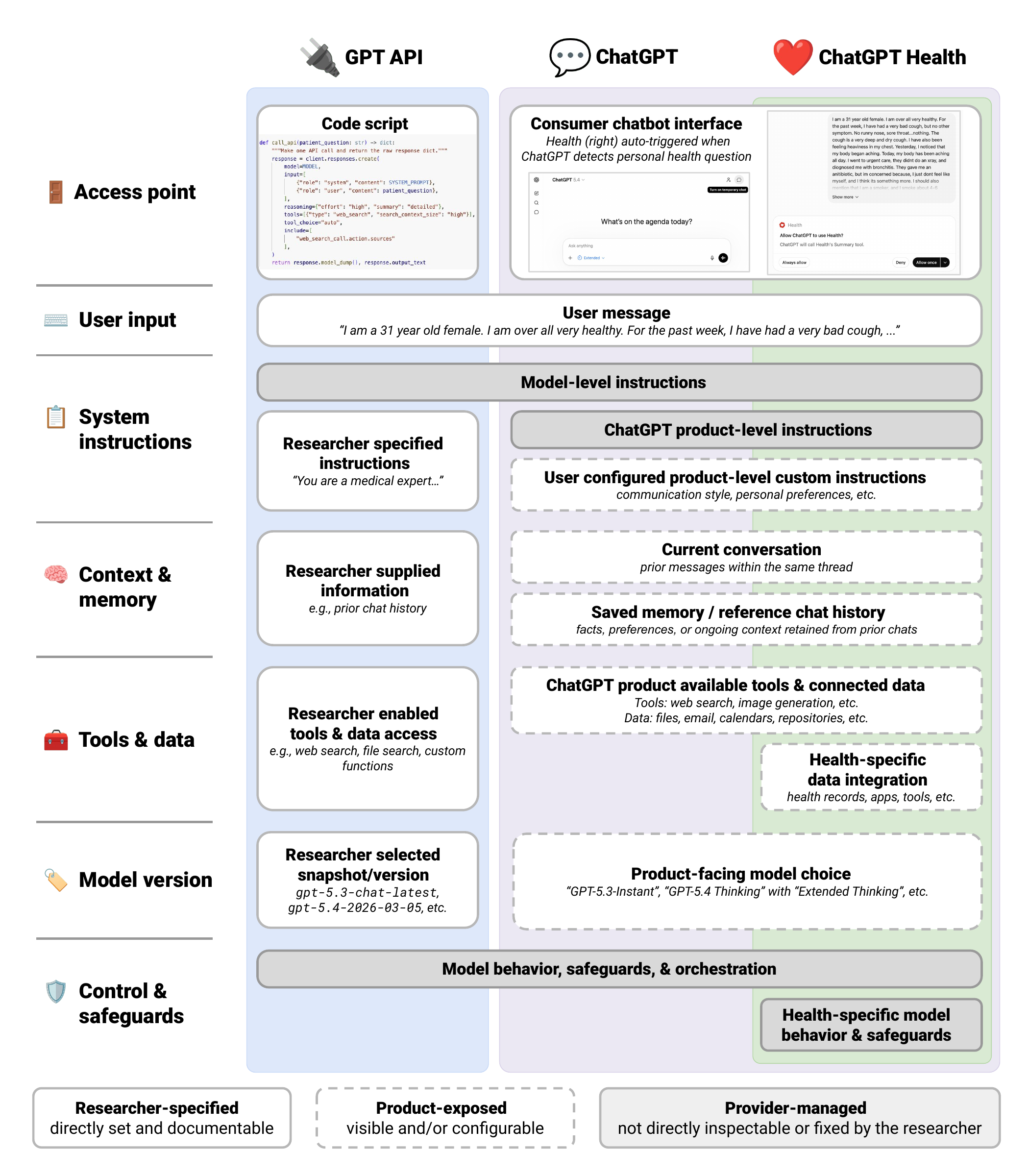}
    \caption{\textbf{Conceptual comparison of researcher-controlled GPT API, ChatGPT, and ChatGPT Health access.} Even for the same user question and model version, the surrounding instructions, context, tools, and safeguards may differ across access modes and influence the response. The schematic is conceptual and does not depict the internal architecture of OpenAI systems. Annotations within boxes provide examples for each component.
}
    \label{fig:access}
\end{figure}

However, the chatbot interface and corresponding API are different access systems. The responses obtained through one are not necessarily equivalent to the other, even when the model version is held constant and available settings are matched as closely as possible (Figure~\ref{fig:access}). For example, the same question submitted to GPT-5.3 through the API and the chatbot interface may be answered differently based on system instructions and memories from prior chat history. Many of these configurations are built into consumer-facing chatbot interface, and not all of them are publicly documented; researchers using the API can configure analogous settings independently but cannot fully replicate the consumer-facing environment. Prior work on AI auditing and governance has emphasized that evaluation conclusions can depend on the form of model access provided, and has called for greater transparency and empirical evaluation across access conditions \cite{casper2024black, kembery2024position}. Nevertheless, findings from either mode are often generalized as if they characterize a single system, although these differences may themselves shape responses and, in turn, influence user perceptions and study conclusions. Empirical comparisons of these access modes remain limited. Outside healthcare, studies have reported differences between ChatGPT interface and GPT API outputs in source selection, source diversity, and behavior in disordered or conspiratorial dialogue \cite{schattoeckrodt2025chatgpt, kirgis2026llm}. 

The recent deployment of Health in ChatGPT\footnote{https://openai.com/index/health-in-chatgpt/, accessed on August 10, 2026} further complicates evaluation within the health domain. 
Health is a dedicated experience within the ChatGPT interface that incorporates health-specific model training, automatically activates when ChatGPT detects a personal health question, and draws on medical records and wellness data, if connected by the user \cite{arora2025healthbench}. 
These add to a growing set of health-specific ChatGPT features that cannot be reproduced through the API (Figure~\ref{fig:access}). 
However, the extent to which the consumer chatbot interface versus the API produces meaningfully different responses to the same health questions remains largely untested.

\section{Study Design}\label{design}

We compared responses to the same consumer health questions across two GPT versions (GPT-5.3 and GPT-5.4) and their available access modes, using fresh sessions with no connected data, conversation history, memory, or user customizations. For GPT-5.3 Instant, the default model for all account holders at the time of data collection, we evaluated three modes of access: standard ChatGPT, ChatGPT Health, and API. For GPT-5.4, which was available to paying subscribers, we compared the standard ChatGPT chatbot interface with the API, with Extended Thinking and web search enabled. For each of 50 patient questions collected from an online medical consultation platform \cite{li2023chatdoctor}, we submitted the question three times in each version-mode condition and used paired within-question comparisons to assess differences in response form, behavior, referenced web sources (for GPT-5.4 only), and clinical content.

\section{Results}\label{results}
\begin{figure}[ht]
    \centering
    \includegraphics[width=\textwidth]{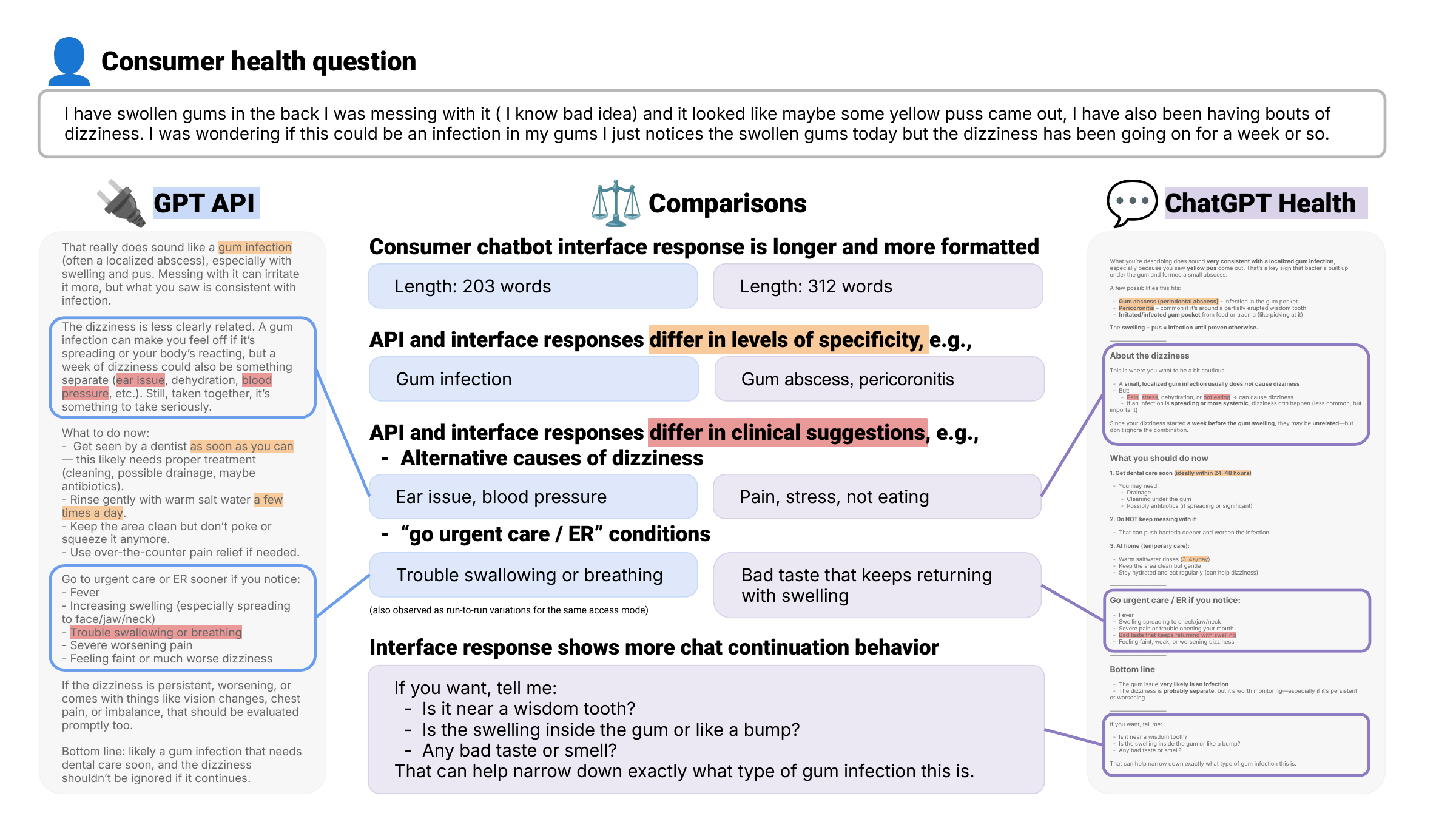}
    \caption{\textbf{Example GPT-5.3 responses to the same consumer health question, accessed through the GPT API and ChatGPT Health interface, annotated for cross-mode differences.} ChatGPT Health is a health-specific experience within ChatGPT that automatically activates for detected personal health questions. Although both access modes used the same nominal model (GPT-5.3), the responses differed in form, behavior, and clinical content.}
    \label{fig:examples}
\end{figure}

Figure~\ref{fig:examples} illustrates how responses to the same health question differed across access modes even within the same model version. These differences were systematic across the 50 questions studied (42 for comparisons involving ChatGPT Health; see Section~\ref{methods:health}) and are detailed below by dimension.

\bmhead{Response form} Responses differed in user-visible form across access modes (Table~\ref{tab:feature_main} and Figure~\ref{fig:examples}). In GPT-5.3, both standard ChatGPT and ChatGPT Health responses were longer, easier to read, and more densely formatted (i.e., more headings, list, emphasis, and emojis) than API responses (paired Wilcoxon tests on question-level means, all $p<0.001$). ChatGPT Health broadly resembled the standard ChatGPT interface answers, although it was easier to read ($p<0.001$), and had slightly more frequently use of section-headings ($p=0.024$). In GPT-5.4, interface responses remained longer than API answers ($p=0.003$), but the main readability and formatting-density differences reversed: interface responses were harder to read and less densely formatted than API responses (all $p<0.001$). Emojis did not appear in responses from either GPT-5.4 access mode.

\bmhead{Response behavior} Access modes also differed in how responses positioned themselves toward the user (Table~\ref{tab:feature_main} and Figure~\ref{fig:examples}). In GPT-5.3, responses from standard ChatGPT and ChatGPT Health invited further interaction much more often than API responses (both $p<0.001$), primarily via more requests for additional information from the user (both $p<0.001$). Conversely, in GPT-5.4, API responses were more likely than ChatGPT responses to make any continuation move ($p<0.001$), including requests for more information ($p<0.001$). Offers to produce additional output differed in the same directions but less consistently, with borderline differences for ChatGPT Health versus standard ChatGPT ($p=0.043$) and GPT-5.4 ChatGPT versus API ($p=0.046$). Self-limiting statements were uncommon in all GPT-5.3 responses regardless of access mode and least frequent in ChatGPT Health, which differed from API responses ($p=0.047$) but not significantly from standard ChatGPT responses ($p=0.250$, only 3 of 42 questions had nonzero paired differences). GPT-5.4 adopted these statements more commonly, particularly in API responses ($p<0.001$).

\bmhead{Referenced sources} When using the web retrieval functionality in GPT 5.4, API and ChatGPT responses cited similar numbers of webpages and domains (Table~\ref{tab:feature_main}), but often different ones. To quantify this variability, we compared the Jaccard overlap\footnote{The proportion of shared items relative to the union of both sets. For example, if two responses each cite six webpages and share only one, the Jaccard similarity is $1/11\approx 9\%$.} within and across access modes. Within a single access mode, repeatedly collected responses to the same question already showed substantial variability, averaging only about 17-18\% for individual webpages and 39\% for domains. Across access modes, Jaccard overlap fell further to 10.9\% for webpages and 29.8\% for domains (both $p<0.001$ versus within-mode overlap).

\bmhead{Clinical content}
We assessed two types of clinical content: the clinical concepts each response mentioned beyond the patient question (e.g., diagnoses, tests, referrals) and the levels of care it named (e.g., self-care, emergency department visit) (see Section~\ref{methods:clinical_content}). Both types varied less by access mode than the response features described above. Applying the same within- versus cross-mode comparison, we found that in GPT-5.3, additional clinical concepts overlapped slightly less between standard ChatGPT and API responses (cross-mode Jaccard of 0.615 v.s. average within-mode Jaccard of 0.653, $p=0.012$). The ChatGPT Health condition showed a similar separation from API responses (0.614 vs. 0.662, $p=0.002$), but was nearly indistinguishable from the standard ChatGPT (0.645 vs. 0.654, $p=0.617$). In GPT-5.4, cross-mode overlap between ChatGPT and API responses was also lower than the averaged within-mode baseline (0.586 vs. 0.619, $p=0.013$). Named levels of care showed the same pattern, with cross-mode overlap close to within-mode overlap. The number of additional clinical concepts differed modestly by access mode: standard GPT-5.3 ChatGPT named slightly more concepts than API responses (6.7 vs. 5.9, $p<0.001$), whereas GPT-5.4 ChatGPT named slightly fewer than API responses (6.7 vs. 7.4, $p=0.005$).

\begin{table}[h]
\caption{Core response features by access mode. For scalar features, the three runs for each question were first averaged within mode, and these question-level means were then summarized across questions. For percentage features, cells summarize the question-level proportion of runs in which the feature was present. API and standard ChatGPT columns include 50 questions; ChatGPT Health includes the 42 questions for which ChatGPT Health activated.}
\label{tab:feature_main}

\footnotesize
\setlength{\tabcolsep}{4pt}

\begin{tabular*}{\textwidth}{@{\extracolsep\fill}lllccccc}
\toprule
& & & \multicolumn{3}{@{}c@{}}{GPT-5.3}
      & \multicolumn{2}{@{}c@{}}{GPT-5.4} \\
\cmidrule{4-6}\cmidrule{7-8}
Dimension & Feature & Measure
  & API & ChatGPT & \makecell{ChatGPT\\Health} & API & ChatGPT \\
\midrule

Sample
  & Questions
  & $n$
  & 50 & 50 & 42 & 50 & 50 \\

\midrule

\multirow{6}{*}{\makecell{Response\\form}}
  & Length
  & words
  & 268 & 377 & 373 & 425 & 451 \\

  & Readability
  & FK grade
  & 8.9 & 8.0 & 7.4 & 10.9 & 11.8 \\

  & Section headings
  & $n$/100 words
  & 0.9 & 2.5 & 2.7 & 0.3 & 0.1 \\

  & List items
  & $n$/100 words
  & 4.1 & 6.4 & 6.6 & 1.7 & 0.4 \\

  & Bold/italic spans
  & $n$/100 words
  & 0.9 & 3.9 & 4.0 & 6.8 & 4.8 \\

  & Any emoji
  & \% of runs
  & 0.7 & 36.7 & 28.6 & 0.0 & 0.0 \\

\midrule

\multirow{4}{*}{\makecell{Response\\behavior}}
  & Any disclaimer
  & \% of runs
  & 10.0 & 6.0 & 0.8 & 42.0 & 24.7 \\

\cmidrule{2-8}

  & Asks for information
  & \% of runs
  & 54.0 & 76.0 & 91.3 & 76.7 & 52.0 \\

  & Offers to produce more
  & \% of runs
  & 24.0 & 35.3 & 23.8 & 26.7 & 19.3 \\

  & Any continuation move
  & \% of runs
  & 70.7 & 98.0 & 100.0 & 98.7 & 67.3 \\

\midrule

\multirow{2}{*}{\makecell{Referenced\\sources}}
  & Webpages referenced
  & $n$
  & --- & --- & --- & 4.7 & 4.4 \\

  & Domains referenced
  & $n$
  & --- & --- & --- & 3.2 & 3.2 \\

\bottomrule
\end{tabular*}
\end{table}

\section{Discussion}
In our study, the largest and most consistent divergences between responses across access modes appeared in readability, formatting, and engagement behavior -- features that may shape patient comprehension and, ultimately, health outcomes \cite{shoemaker2014development, sheridan2011interventions}. Several differences also reversed direction when the underlying GPT model changed, indicating that access-mode effects are not stable across model versions. 

Our findings in health are consistent with contemporaneous reports of interface--API differences in other settings \cite{encarnacion2026what, sadowski2025scraped}. Together, these results show that valid auditing of AI chatbots requires not only realistic questions and rigorous scoring, but also a way to scalably recreate the settings through which consumers actually encounter the system. Currently, these configurations are not publicly documented or accessible in sufficient detail for independent evaluation.

Importantly, our study examined access modes under unusually controlled conditions and still found significant differences: single-turn questions without memory, without connected records, without conversation history, and without user customization. In routine consumer use, these features substantially alter model behavior and could further widen the gap between what consumers observe and what API-based evaluations can observe. A recent lawsuit illustrates the stakes: ChatGPT allegedly customized health-related responses based on a user's religious identity, and didn't recommend immediate care when the user was suffering from a pulmonary embolism \cite{seddon2026chatgpt}. Being able to recreate such scenarios is central to evaluating a model's safety, but these customizations aren't reproducible through the standard available channels.

This inability to replicate consumer experiences creates an important regulatory blind spot, that cannot be fixed by researchers on their own. Regulators should require developers of consumer-facing health AI to provide APIs that provide researchers with scalable and reproducible access to consumer deployment pathways. As is, there are no ways for independent researchers to scalably audit minor feature differences (e.g., access mode), much less the major features that are rolling out (e.g. connection to the EHR). Without such requirements, independent evaluation risks measuring the model available to auditors rather than the product experienced by patients. Otherwise, we will be effectively ceding all assessment of consumer health AI to the frontier model providers themselves.

\newpage
\section{Methods}\label{method}

\subsection{Consumer health question collection}
We sampled 50 patient-authored consumer health questions from \texttt{HealthCareMagic-100k} \cite{li2023chatdoctor}, a corpus of online medical questions and physician responses. To better approximate consumer health questions submitted to a general-purpose chatbot, we restricted the eligible question pool before sampling to queries without explicit addressee framing, excluding questions that directly addressed the respondent as a physician or clinician through terms such as ``doctor''. Only the patient question text was used in this study; physician answers and other response-side information were not provided to the models.

\subsection{LLM response collection}
We evaluated GPT responses across the ChatGPT web interface and the OpenAI API, with an additional condition using Health in ChatGPT. The primary comparison included two GPT versions across two access modes: GPT-5.3 Instant, the default model for all account holders at the time of collection, and GPT-5.4 Thinking with Extended Thinking and web search enabled, available to paying subscribers.\footnote{https://help.openai.com/en/articles/11909943-gpt-53-and-gpt-54-in-chatgpt, accessed on April 12, 2026} Each of the 50 patient questions was submitted in three independent sessions per version and mode, yielding 600 unique responses in total: 2 versions $\times$ 2 access modes $\times$ 50 questions $\times$ 3 runs. A subsequent Health condition used GPT-5.3 Instant and yielded eligible Health responses for 42 questions, resulting in 126 additional responses. Repeated runs were collected to characterize run-to-run variability in responses generated for the same question under the same nominal condition. 

For the response collection, queries were administered as a naive user would encounter the system: each question was submitted verbatim as the sole input, with no personalization, follow-up turns, or task-specific customized instructions. Data for the primary comparisons were collected between April 1st and 22nd, 2026, after the most recent update to GPT-5.3 Instant (March 16th, 2026)\footnote{https://help.openai.com/en/articles/9624314-model-release-notes, accessed on April 12, 2026} and before the release of the successor GPT-5.5 family (April 23rd, 2026)\footnote{https://openai.com/index/introducing-gpt-5-5/, accessed on June 6, 2026}. The Health condition was collected separately on August 4th to 5th, 2026, following the broader U.S. rollout of Health in ChatGPT in July. 

\bmhead{ChatGPT Interface} For the ChatGPT interface condition, each run began in a new conversation to avoid carryover from prior turns. Interface queries were conducted in Temporary Chat mode, which excludes the exchange from ChatGPT's persistent cross-conversation memory.\footnote{https://help.openai.com/en/articles/8914046-temporary-chat-faq, accessed on April 12, 2026} GPT-5.3 Instant responses were captured using the interface copy function. For GPT-5.4 Thinking, the copy function exported only the final answer. Because the rendered interface also displayed a brief pre-answer preamble before the final response, we manually selected the full rendered output for GPT-5.4 interface responses.

\bmhead{ChatGPT Health} Health in ChatGPT is a health-focused experience within the ChatGPT consumer interface first introduced in January 2026, with a broader U.S. rollout beginning on July 23, 2026. OpenAI reports that models supporting health conversations receive dedicated health-focused training and are developed and evaluated with physician input, including physician-authored evaluation rubrics \cite{arora2025healthbench}. Users can also optionally connect supported medical records and health or wellness data. At the time of response collection, GPT-5.3 Instant was the only one of the two primary model settings still available in ChatGPT; responses were therefore collected using this setting with web search disabled. OpenAI’s public release notes documented no intervening update to GPT-5.3 Instant between the April and August collection periods. For this collection, no medical records or health applications were connected. As Temporary Chat was not available in Health during data collection, each question was submitted in a separate conversation and deleted immediately after collection to minimize cross-conversation memory effects. Health was manually selected for all 50 questions but activated for only 42. The remaining eight, all of which concerned another person’s health rather than the user’s own, did not activate Health and were therefore excluded from the Health analyses. \label{methods:health}

\bmhead{API} For the API condition, we queried the corresponding model versions with an empty system prompt and otherwise used settings chosen to mirror the observed interface behavior. GPT-5.3 Instant was accessed as \texttt{gpt-5.3-chat-latest} with no reasoning mode and no tools. GPT-5.4 Thinking was accessed as \texttt{gpt-5.4} (the only available snapshot was \texttt{gpt-5.4-2026-03-05}) with high reasoning effort and web search enabled (\texttt{tools=[\{"type": "web\_search", "search\_context\_size": "high"\}], tool\_choice="auto"}). Other parameters were left at their defaults. 

\subsection{LLM response processing}
For all responses regardless of version, we removed Markdown syntax before computing length and readability measures so that these reflected prose rather than formatting. Formatting features were measured before Markdown removal. For GPT-5.4 responses, web reference related formatting was first separated from response prose before other feature extraction. Cited links were retained separately for source-reference analyses. In all GPT-5.4 ChatGPT responses, the interface displayed a brief preamble before the final answer that signaled the model's intended approach; the preamble was not present in responses collected from the API. Our primary analysis used the full user-visible response including the brief pre-answer preamble; final-answer-only length related analyses are reported in the Appendix.

\subsection{Response features}
\bmhead{Response form}
Using the markdown-free response text, response \textit{length} was measured as the number of words and \textit{readability} as the Flesch-Kincaid grade level. Formatting features captured visible response structure. \textit{Section headings} included Markdown headings, standalone bold heading lines, and short colon-terminated labels. \textit{List items} included ordered and unordered list lines. \textit{Bold/italic spans} were counted as inline emphasis, excluding standalone bold section-heading lines to avoid double-counting heading structure. \textit{Emoji} was identified using Unicode.

\bmhead{Response behavior} \textit{Explicit self-limiting disclaimers} were measured as a binary indicator of whether a response contained at least one sentence in which the model directly limited its own role or capacity using the phrases ``I can't'' or ``I cannot'', identified using regular expressions. This feature was intended to capture a narrow, reproducible form of self-limiting language rather than the broader set of medical disclaimers, safety statements, or care-escalation advice. \textit{Continuation moves} were coded as requests for more information, offers to produce additional output, or either type of move. Requests for more information included questions or prompts for additional case details. Offers to produce additional output included offers to generate a checklist, message draft, plan, summary, or similar follow-up material without requiring new case facts. These two categories were not mutually exclusive. Continuation moves were classified by GPT-5.4 using the definitions above.

\bmhead{Referenced sources} For GPT-5.4 responses, \textit{webpages referenced} were extracted from the original response text and deduplicated within each response. Each cited webpage was also mapped to a source-domain label, with recurring subdomain variants collapsed to the same parent source domain, together forming \textit{domain referenced}.

\bmhead{Clinical content} \label{methods:clinical_content}
Clinical concepts were defined as distinct clinical conditions, tests or workups, treatments, or specialist referrals. We focused on \textit{additional clinical concepts}: concepts present in a response but not already raised in the patient question. All extraction was performed using a GPT-5.4-based pipeline with clinician-reviewed extraction prompts iteratively refined against extraction results. For each sampled question, we constructed a question-specific codebook of clinical concepts from the patient question and all responses. We then annotated the patient question and each response against the same codebook for concept presence or absence. Additional clinical concepts included diagnoses, specialist referrals, clinical treatments, and clinical tests or workups. \textit{Named levels of care} were extracted separately from the clinical-concept codebook using a fixed-label schema: self-care, routine follow-up, urgent follow-up, emergency department care, and calling 911.

\subsection{Statistical analysis} 
We used the question as the primary unit of comparison in this study. The three repeated runs for a given question, model version, and access mode were used to estimate that condition's behavior for that question. All access-mode comparisons were therefore paired within question and performed separately for GPT-5.3 and GPT-5.4.

For scalar response-level features, including length, readability, formatting-feature densities, number of unique cited webpages, number of unique cited source domains, number of additional clinical concepts, we first calculated the feature for each individual response and then averaged across the three runs within each question, model version, and access mode. Access-mode values were then compared within the same question using paired Wilcoxon signed-rank tests.

For binary response-level features, including emoji use, explicit self-limiting statements, and continuation moves, we calculated the proportion of the three runs in which the feature appeared for each question, model version, and access mode. These question-level proportions could take values of 0/3, 1/3, 2/3, or 3/3 and were compared between access modes using paired Wilcoxon signed-rank tests when there were sufficient nonzero paired differences. Response-level percentages were also reported descriptively to aid interpretation.

For set-valued response features, we quantified consistency using Jaccard overlap. These features represented each response as a set of extracted items, such as cited webpages, cited source domains, additional clinical concepts, and named levels of care. Within-mode consistency was computed for each question by averaging the Jaccard values of the three pairwise comparisons among repeated runs in the same access mode. Cross-mode consistency was computed analogously by averaging the Jaccard values of all nine run pairs formed by crossing the three runs from each of the two access modes being compared. Both measures operated at the individual-run level rather than pooling items across runs before comparison. Jaccard overlap was undefined and excluded when the union of the compared item sets was empty; analytic sample sizes therefore varied by feature and access-mode pair. To test whether access mode introduced systematic differences beyond run-to-run variability, we compared each question's cross-mode Jaccard overlap with the mean of its two within-mode overlaps using a paired Wilcoxon signed-rank test.

Complete descriptive statistics, paired feature contrasts, and within- versus cross-mode overlap contrasts are reported in Tables~\ref{tab:feature_full}, \ref{tab:form_pvalues}, and \ref{tab:overlap_contrasts}, respectively.

\backmatter

\section*{Declarations}

\bmhead{Data availability} Processed study materials, including the sampled health questions and collected model responses, will be available at https://github.com/yuan-pu/health-llm-variability, subject to applicable source-data terms.

\bmhead{Code availability} Code and prompts used to process responses, extract study features, and reproduce the reported analyses will be available at https://github.com/yuan-pu/health-llm-variability.

\bmhead{Competing interests}
The authors declare no competing interests.

\bigskip

\bibliography{sn-bibliography}

@online{montero2026kff,
  author       = {Montero, Alex and Montalvo, III, Julian and Kearney, Audrey and Valdes, Isabelle and Kirzinger, Ashley and Hamel, Liz},
  title        = {{KFF} Tracking Poll on Health Information and Trust: Use of {AI} for Health Information and Advice},
  organization = {KFF},
  date         = {2026-03-25},
  url          = {https://www.kff.org/public-opinion/kff-tracking-poll-on-health-information-and-trust-use-of-ai-for-health-information-and-advice/},
  urldate      = {2026-06-09}
}

@online{holzwarth2026tortoise,
  author       = {Holzwarth, Mikayla and Nagappan, Ashwini and Knowles, Madelyn},
  title        = {The Tortoise and the Hare of Care: Health {AI} Insights from {Rock Health's} 2025 {Consumer Adoption Survey}},
  organization = {Rock Health},
  date         = {2026-03-23},
  url          = {https://rockhealth.com/insights/the-tortoise-and-the-hare-of-care-health-ai-insights-from-rock-healths-2025-consumer-adoption-survey/},
  urldate      = {2026-06-09}
}

@article{ayers2023comparing,
    author = {Ayers, John W. and Poliak, Adam and Dredze, Mark and Leas, Eric C. and Zhu, Zechariah and Kelley, Jessica B. and Faix, Dennis J. and Goodman, Aaron M. and Longhurst, Christopher A. and Hogarth, Michael and Smith, Davey M.},
    title = {Comparing Physician and Artificial Intelligence Chatbot Responses to Patient Questions Posted to a Public Social Media Forum},
    journal = {JAMA Internal Medicine},
    volume = {183},
    number = {6},
    pages = {589-596},
    year = {2023},
    month = {06},
    issn = {2168-6106},
    doi = {10.1001/jamainternmed.2023.1838},
    url = {https://doi.org/10.1001/jamainternmed.2023.1838},
}

@article{pan2023assessment,
    author = {Pan, Alexander and Musheyev, David and Bockelman, Daniel and Loeb, Stacy and Kabarriti, Abdo E.},
    title = {Assessment of Artificial Intelligence Chatbot Responses to Top Searched Queries About Cancer},
    journal = {JAMA Oncology},
    volume = {9},
    number = {10},
    pages = {1437-1440},
    year = {2023},
    month = {10},
    issn = {2374-2437},
    doi = {10.1001/jamaoncol.2023.2947},
    url = {https://doi.org/10.1001/jamaoncol.2023.2947},
}

@article{musheyev2024readability,
    author = {Musheyev, David and Pan, Alexander and Gross, Preston and Kamyab, Daniel and Kaplinsky, Peter and Spivak, Mark and Bragg, Marie A. and Loeb, Stacy and Kabarriti, Abdo E.},
    title = {Readability and Information Quality in Cancer Information From a Free vs Paid Chatbot},
    journal = {JAMA Network Open},
    volume = {7},
    number = {7},
    pages = {e2422275-e2422275},
    year = {2024},
    month = {07},
    issn = {2574-3805},
    doi = {10.1001/jamanetworkopen.2024.22275},
    url = {https://doi.org/10.1001/jamanetworkopen.2024.22275},
}

@article{draelos2026large,
    title={Large language models provide unsafe answers to patient-posed medical questions},
author={Draelos, Rachel L. and Afreen, Samina and Blasko, Barbara and Brazile, Tiffany L. and Chase, Natasha and Desai, Dimple Patel and Evert, Jessica and Gardner, Heather L. and Herrmann, Lauren and House, Aswathy Vaikom and Kass, Stephanie and Kavan, Marianne and Khemani, Kirshma and Koire, Amanda and McDonald, Lauren M. and Rabeeah, Zahraa and Shah, Amy},
    journal={npj Digital Medicine},
    volume={9},
    pages={241},
    year={2026},
    doi={10.1038/s41746-026-02428-5},
    publisher={Nature Publishing Group UK London}
}

@misc{jin2024better,
author = {Jin, Yiqiao and Chandra, Mohit and Verma, Gaurav and Hu, Yibo and De Choudhury, Munmun and Kumar, Srijan},
title = {Better to Ask in English: Cross-Lingual Evaluation of Large Language Models for Healthcare Queries},
year = {2024},
isbn = {9798400701719},
publisher = {Association for Computing Machinery},
address = {New York, NY, USA},
doi = {10.1145/3589334.3645643},
booktitle = {Proceedings of the ACM Web Conference 2024},
pages = {2627–2638},
numpages = {12},
location = {Singapore, Singapore},
series = {WWW '24}
}

@article{abrar2025empirical,
AUTHOR = {Abrar, Moaiz and Sermet, Yusuf and Demir, Ibrahim},
TITLE = {An Empirical Evaluation of Large Language Models on Consumer Health Questions},
JOURNAL = {BioMedInformatics},
VOLUME = {5},
YEAR = {2025},
NUMBER = {1},
ARTICLE-NUMBER = {12},
URL = {https://www.mdpi.com/2673-7426/5/1/12},
ISSN = {2673-7426},
DOI = {10.3390/biomedinformatics5010012}
}

@article{fernandez2025evaluating,
  title={Evaluating search engines and large language models for answering health questions},
  author={Fern{\'a}ndez-Pichel, Marcos and Pichel, Juan C and Losada, David E},
  journal={npj Digital Medicine},
  volume={8},
  pages={153},
  year={2025},
  doi={10.1038/s41746-025-01546-w},
  publisher={Nature Publishing Group UK London}
}

@article{sharma2025longitudinal,
  title={A longitudinal analysis of declining medical safety messaging in generative AI models},
  author={Sharma, Sonali and Alaa, Ahmed M and Daneshjou, Roxana},
  journal={npj Digital Medicine},
  volume={8},
  pages={592},
  year={2025},
  doi={10.1038/s41746-025-01943-1},
  publisher={Nature Publishing Group UK London}
}

@article{kopka2026evaluating,
  title={Evaluating the accuracy of {ChatGPT} model versions for giving care-seeking advice},
  author={Kopka, Marvin and He, Longqi and Feufel, Markus A},
  journal={Communications Medicine},
  volume={6},
  pages={171},
  year={2026},
  doi={10.1038/s43856-026-01466-0},
  publisher={Nature Publishing Group UK London}
}

@misc{schattoeckrodt2025chatgpt, 
    title={{ChatGPT} as a news recommender system: Measuring source types and diversity across different interfaces}, 
    url={osf.io/preprints/socarxiv/wjzp3_v3}, 
    DOI={10.31235/osf.io/wjzp3_v3}, 
    publisher={SocArXiv}, 
    author={Schatto-Eckrodt, Tim and Liebig, Laura and Reiss, Michael V. and Geislinger, Robert and Schaetz, Nadja and Merten, Lisa and Schröder, Justin T and Königslöw, Katharina K and Laugwitz, Laura and Knor, Eva L and et al.}, 
    year={2025}, 
    month={Nov} 
}

@misc{kirgis2026llm,
      title={LLM Spirals of Delusion: A Benchmarking Audit Study of {AI} Chatbot Interfaces}, 
      author={Peter Kirgis and Ben Hawriluk and Sherrie Feng and Aslan Bilimer and Sam Paech and Zeynep Tufekci},
      year={2026},
      eprint={2604.06188},
      archivePrefix={arXiv},
      primaryClass={cs.HC},
      url={https://arxiv.org/abs/2604.06188}, 
}

@article{li2023chatdoctor,
  title={Chatdoctor: A medical chat model fine-tuned on a large language model meta-ai (llama) using medical domain knowledge},
  author={Li, Yunxiang and Li, Zihan and Zhang, Kai and Dan, Ruilong and Jiang, Steve and Zhang, You},
  journal={Cureus},
  volume={15},
  number={6},
  year={2023},
  publisher={Cureus},
  doi={10.7759/cureus.40895}
}

@misc{arora2025healthbench,
      title={HealthBench: Evaluating Large Language Models Towards Improved Human Health}, 
      author={Rahul K. Arora and Jason Wei and Rebecca Soskin Hicks and Preston Bowman and Joaquin Quiñonero-Candela and Foivos Tsimpourlas and Michael Sharman and Meghan Shah and Andrea Vallone and Alex Beutel and Johannes Heidecke and Karan Singhal},
      year={2025},
      eprint={2505.08775},
      archivePrefix={arXiv},
      primaryClass={cs.CL},
      url={https://arxiv.org/abs/2505.08775}, 
}

@article{shoemaker2014development,
title = {Development of the Patient Education Materials Assessment Tool (PEMAT): A new measure of understandability and actionability for print and audiovisual patient information},
journal = {Patient Education and Counseling},
volume = {96},
number = {3},
pages = {395-403},
year = {2014},
note = {Communication in Healthcare: Lessons from Diversity},
issn = {0738-3991},
doi = {https://doi.org/10.1016/j.pec.2014.05.027},
url = {https://www.sciencedirect.com/science/article/pii/S073839911400233X},
author = {Sarah J. Shoemaker and Michael S. Wolf and Cindy Brach}
}

@article{sheridan2011interventions,
author = {Stacey L. Sheridan and David J. Halpern and Anthony J. Viera and Nancy D. Berkman and Katrina E. Donahue and Karen Crotty},
title = {Interventions for Individuals with Low Health Literacy: A Systematic Review},
journal = {Journal of Health Communication},
volume = {16},
number = {sup3},
pages = {30--54},
year = {2011},
publisher = {Taylor \& Francis},
doi = {10.1080/10810730.2011.604391},
note ={PMID: 21951242},
URL = {https://doi.org/10.1080/10810730.2011.604391}
}

@misc{casper2024black,
author = {Casper, Stephen and Ezell, Carson and Siegmann, Charlotte and Kolt, Noam and Curtis, Taylor Lynn and Bucknall, Benjamin and Haupt, Andreas and Wei, Kevin and Scheurer, J{\'e}r{\'e}my and Hobbhahn, Marius and Sharkey, Lee and Krishna, Satyapriya and Von Hagen, Marvin and Alberti, Silas and Chan, Alan and Sun, Qinyi and Gerovitch, Michael and Bau, David and Tegmark, Max and Krueger, David and Hadfield-Menell, Dylan},
title = {Black-Box Access is Insufficient for Rigorous {AI} Audits},
year = {2024},
isbn = {9798400704505},
publisher = {Association for Computing Machinery},
address = {New York, NY, USA},
doi = {10.1145/3630106.3659037},
booktitle = {Proceedings of the 2024 ACM Conference on Fairness, Accountability, and Transparency},
pages = {2254–2272},
numpages = {19},
location = {Rio de Janeiro, Brazil},
series = {FAccT '24}
}

@misc{kembery2024position,
title={Position Paper: Model Access should be a Key Concern in AI Governance},
author={Edward Kembery},
booktitle={Workshop on Socially Responsible Language Modelling Research},
year={2024},
url={https://openreview.net/forum?id=3OChrbgcMG}
}

@misc{encarnacion2026what,
      title={What Current {AI} Benchmarks Leave Unmeasured: Modality, Search, Citations, and Implications (for Safety Evaluations)}, 
      author={Ro Encarnación and Tina Behzad and Emma Lurie and Danaé Metaxa},
      year={2026},
      eprint={2608.06202},
      archivePrefix={arXiv},
      primaryClass={cs.HC},
      url={https://arxiv.org/abs/2608.06202}, 
}

@online{sadowski2025scraped,
  author  = {Sadowski, Jakub},
  title   = {Scraped {AI} Answers vs. {API} Results from {LLMs}. Is There a Difference? [AI Search Study]},
  year    = {2025},
  date    = {2025-12-05},
  url     = {https://surferseo.com/blog/llm-scraped-ai-answers-vs-api-results/},
  urldate = {2026-06-09},
  organization = {Surfer}
}

@online{seddon2026chatgpt,
  author  = {Seddon, Sean},
  title   = {{ChatGPT} Medical Advice Brought Man ``to Brink of Death'', Lawsuit Alleges},
  year    = {2026},
  date    = {2026-07-23},
  url     = {https://www.bbc.com/news/articles/cwylp3nxp5yo},
  urldate = {2026-09-02},
  organization = {BBC News}
}

\newpage
\begin{appendices}

\section{Additional Tables}\label{additional_tables}

\begin{sidewaystable}[h]
\caption{Full response features and within-mode set overlap by access mode. Cells report mean (standard deviation) across questions. For scalar features, the three runs for each question were first averaged within mode, and these question-level means were then summarized across questions. For percentage features, cells summarize the question-level proportion of runs in which the feature was present. Within-mode overlap rows report mean Jaccard overlap among the three pairwise comparisons of repeated runs within a question and access mode. API and standard ChatGPT columns include 50 questions; ChatGPT Health includes the 42 questions for which Health activated. For within-mode overlap rows, questions with no coded items in any of the compared runs yield undefined Jaccard values and are excluded; the contributing numbers for ``additional clinical concepts'' are 49, 49, 41, 49, and 50, and for ``named levels of care'' are 48, 48, 41, 48, and 49, respectively, in column order.}
\label{tab:feature_full}

\footnotesize
\setlength{\tabcolsep}{3.5pt}
\renewcommand{\arraystretch}{0.96}

\begin{tabular*}{\textwidth}{@{\extracolsep\fill}lllccccc}
\toprule
& & & \multicolumn{3}{@{}c@{}}{GPT-5.3}
      & \multicolumn{2}{@{}c@{}}{GPT-5.4} \\
\cmidrule{4-6}\cmidrule{7-8}
Dimension & Feature & Measure
  & API & ChatGPT & \makecell{ChatGPT\\Health} & API & ChatGPT \\
\midrule

Sample
  & Questions
  & $n$
  & 50 & 50 & 42 & 50 & 50 \\

\midrule

\multirow{11}{*}{\makecell{Response\\form}}

  & Length
  & words
  & 268 (63) & 377 (78) & 373 (82) & 425 (113) & 451 (112) \\

  & Length without preamble
  & words
  & --- & --- & --- & 425 (113) & 380 (103) \\

  & Readability
  & FK grade
  & 8.9 (1.2) & 8.0 (1.2) & 7.4 (0.9) & 10.9 (1.6) & 11.8 (1.6) \\

  & Paragraph length
  & words
  & 27.0 (4.2) & 21.6 (5.1) & 20.0 (3.7) & 52.1 (7.5) & 49.4 (7.2) \\

  & \multirow{2}{*}{Section headings}
  & any, \% of runs
  & 92.0 (18.5) & 99.3 (4.7) & 100.0 (0.0) & 77.3 (27.3) & 34.0 (34.7) \\
  &
  & $n$/100 words
  & 0.9 (0.5) & 2.5 (0.9) & 2.7 (0.7) & 0.3 (0.2) & 0.1 (0.1) \\

  & \multirow{2}{*}{List items}
  & any, \% of runs
  & 98.7 (6.6) & 100.0 (0.0) & 100.0 (0.0) & 84.0 (23.6) & 36.0 (33.6) \\
  &
  & $n$/100 words
  & 4.1 (1.4) & 6.4 (1.9) & 6.6 (1.3) & 1.7 (0.9) & 0.4 (0.6) \\

  & \multirow{2}{*}{Bold/italic spans}
  & any, \% of runs
  & 42.7 (31.6) & 100.0 (0.0) & 100.0 (0.0) & 99.3 (4.7) & 98.7 (6.6) \\
  &
  & $n$/100 words
  & 0.9 (1.0) & 3.9 (1.0) & 4.0 (1.1) & 6.8 (1.0) & 4.8 (1.2) \\

  & Emoji
  & any, \% of runs
  & 0.7 (4.7) & 36.7 (37.0) & 28.6 (35.0) & 0.0 (0.0) & 0.0 (0.0) \\

\midrule

\multirow{4}{*}{\makecell{Response\\behavior}}

  & Disclaimer
  & any, \% of runs
  & 10.0 (27.1) & 6.0 (21.0) & 0.8 (5.1) & 42.0 (36.8) & 24.7 (31.5) \\

\cmidrule{2-8}

  & Asks for information
  & \% of runs
  & 54.0 (35.6) & 76.0 (33.0) & 91.3 (22.2) & 76.7 (33.8) & 52.0 (37.0) \\

  & Offers to produce more
  & \% of runs
  & 24.0 (34.4) & 35.3 (41.7) & 23.8 (37.0) & 26.7 (37.5) & 19.3 (27.8) \\

  & Any continuation move
  & \% of runs
  & 70.7 (32.0) & 98.0 (10.5) & 100.0 (0.0) & 98.7 (6.6) & 67.3 (31.9) \\

\midrule

\multirow{4}{*}{\makecell{Referenced\\sources}}

  & \multirow{2}{*}{Webpages referenced}
  & $n$
  & --- & --- & --- & 4.7 (1.4) & 4.4 (1.1) \\
  &
  & within-mode overlap
  & --- & --- & --- & 0.173 (0.121) & 0.177 (0.118) \\

  & \multirow{2}{*}{Domains referenced}
  & $n$
  & --- & --- & --- & 3.2 (1.0) & 3.2 (0.7) \\
  &
  & within-mode overlap
  & --- & --- & --- & 0.391 (0.207) & 0.386 (0.179) \\

\midrule

\multirow{4}{*}{\makecell{Clinical\\content}}

  & \multirow{2}{*}{Additional clinical concepts}
  & $n$
  & 5.9 (3.4) & 6.7 (3.8) & 6.0 (3.7) & 7.4 (4.7) & 6.7 (4.1) \\
  &
  & within-mode overlap
  & 0.658 (0.181) & 0.647 (0.176) & 0.651 (0.141)
  & 0.636 (0.184) & 0.590 (0.192) \\

  & \multirow{2}{*}{Named levels of care}
  & $n$
  & 3.0 (1.4) & 2.8 (1.1) & 2.8 (1.3) & 4.2 (1.5) & 4.1 (1.8) \\
  &
  & within-mode overlap
  & 0.687 (0.244) & 0.696 (0.275) & 0.725 (0.260)
  & 0.727 (0.187) & 0.745 (0.185) \\

\bottomrule
\end{tabular*}
\end{sidewaystable}

\begin{sidewaystable}[h]
\caption{Paired access-mode contrasts for response features. Cells report $\Delta$ ($p$), where $\Delta$ is the mean paired difference and $p$ is the unadjusted p-value from a two-sided paired Wilcoxon signed-rank test. Tests use question-level means for scalar features and question-level feature prevalence for percentage features; for percentage features, $\Delta$ is in percentage points. Positive $\Delta$ values favor the first-named mode in each column heading. A=API, C=ChatGPT, H=ChatGPT Health. For Flesch--Kincaid grade, lower values indicate easier readability. ``n.t.'' indicates contrasts not tested because fewer than two questions produced a nonzero paired difference; dashes indicate contrasts absent from the design. Contrasts involving ChatGPT Health use the matched 42-question subset; therefore, their $\Delta$ values need not equal differences between the corresponding descriptive column means, for which API and standard ChatGPT summarize all 50 questions.}
\label{tab:form_pvalues}

\footnotesize
\setlength{\tabcolsep}{3.5pt}
\renewcommand{\arraystretch}{0.96}

\begin{tabular*}{\textwidth}{@{\extracolsep\fill}lllcccc}
\toprule
& & & \multicolumn{3}{@{}c@{}}{GPT-5.3}
      & \multicolumn{1}{@{}c@{}}{GPT-5.4} \\
\cmidrule{4-6}\cmidrule{7-7}
Dimension & Feature & Measure
  & C vs A & H vs A & H vs C & C vs A \\
\midrule

\multirow{10}{*}{\makecell{Response\\form}}

  & Length
  & words
  & +109 ($<$0.001)
  & +103 ($<$0.001)
  & $-$8 (0.132)
  & +26 (0.003) \\

  & Length without preamble
  & words
  & ---
  & ---
  & ---
  & $-$45 ($<$0.001) \\

  & Readability
  & FK grade
  & $-$0.89 ($<$0.001)
  & $-$1.38 ($<$0.001)
  & $-$0.54 ($<$0.001)
  & +0.90 ($<$0.001) \\

  & \multirow{2}{*}{Section headings}
  & any, \% of runs
  & +7.3 (0.004)
  & +7.9 (0.008)
  & n.t.
  & $-$43.3 ($<$0.001) \\

  &
  & $n$/100 words
  & +1.56 ($<$0.001)
  & +1.82 ($<$0.001)
  & +0.28 (0.024)
  & $-$0.19 ($<$0.001) \\

  & \multirow{2}{*}{List items}
  & any, \% of runs
  & +1.3 (0.500)
  & +1.6 (0.500)
  & n.t.
  & $-$48.0 ($<$0.001) \\

  &
  & $n$/100 words
  & +2.29 ($<$0.001)
  & +2.69 ($<$0.001)
  & +0.39 (0.060)
  & $-$1.25 ($<$0.001) \\

  & \multirow{2}{*}{Bold/italic spans}
  & any, \% of runs
  & +57.3 ($<$0.001)
  & +57.1 ($<$0.001)
  & n.t.
  & $-$0.7 (1.000) \\

  &
  & $n$/100 words
  & +2.93 ($<$0.001)
  & +3.03 ($<$0.001)
  & +0.20 (0.171)
  & $-$2.09 ($<$0.001) \\

  & Emoji
  & any, \% of runs
  & +36.0 ($<$0.001)
  & +28.6 ($<$0.001)
  & $-$6.3 (0.336)
  & n.t. \\

\midrule

\multirow{2}{*}{\makecell{Referenced\\sources}}

  & Webpages referenced
  & $n$
  & ---
  & ---
  & ---
  & $-$0.27 (0.109) \\

  & Domains referenced
  & $n$
  & ---
  & ---
  & ---
  & $-$0.04 (0.820) \\

\midrule

\multirow{4}{*}{\makecell{Response\\ behavior}}

  & Disclaimer
  & any, \% of runs
  & $-$4.0 (0.391)
  & $-$10.3 (0.047)
  & $-$4.8 (0.250)
  & $-$17.3 ($<$0.001) \\

\cmidrule{2-7}

  & Asks for information
  & \% of runs
  & +22.0 ($<$0.001)
  & +31.0 ($<$0.001)
  & +11.1 ($<$0.001)
  & $-$24.7 ($<$0.001) \\

  & Offers to produce more
  & \% of runs
  & +11.3 (0.014)
  & $-$0.8 (0.861)
  & $-$10.3 (0.043)
  & $-$7.3 (0.046) \\

  & Any continuation move
  & \% of runs
  & +27.3 ($<$0.001)
  & +23.8 ($<$0.001)
  & n.t.
  & $-$31.3 ($<$0.001) \\

\midrule

\multirow{2}{*}{\makecell{Clinical\\content}}

  & Additional clinical concepts
  & $n$
  & +0.73 ($<$0.001)
  & +0.09 (0.900)
  & $-$0.56 ($<$0.001)
  & $-$0.73 (0.005) \\

  & Named levels of care
  & $n$
  & $-$0.23 (0.213)
  & $-$0.29 (0.333)
  & +0.05 (0.693)
  & $-$0.11 (0.627) \\

\bottomrule
\end{tabular*}
\end{sidewaystable}

\begin{table}[h]
\caption{Within- versus cross-mode set overlap. $n$ is the number of questions contributing both a cross-mode and within-mode overlap for the access-mode pair. Cross is the mean question-level Jaccard overlap across the access-mode pair. Within is the average of the two within-mode repeated-run Jaccard overlaps for the same questions. $\Delta$ is within minus cross. P-values are unadjusted values from two-sided paired Wilcoxon signed-rank tests on question-level Jaccard overlaps. A=API, C=ChatGPT, H=ChatGPT Health.}
\label{tab:overlap_contrasts}
\footnotesize
\setlength{\tabcolsep}{4pt}
\begin{tabular*}{\textwidth}{@{\extracolsep\fill}llccccc}
\toprule
Metric & Access pair & $n$ & Cross & Within & $\Delta$ & p-value \\
\midrule
Referenced webpage overlap & GPT-5.4 C/A & 50 & 0.109 & 0.175 & +0.066 & $<$0.001 \\
Referenced domain overlap & GPT-5.4 C/A & 50 & 0.298 & 0.389 & +0.091 & $<$0.001 \\
\midrule
Additional clinical concept overlap & GPT-5.3 C/A & 49 & 0.615 & 0.653 & +0.038 & 0.012 \\
Additional clinical concept overlap & GPT-5.3 H/A & 41 & 0.614 & 0.662 & +0.048 & 0.002 \\
Additional clinical concept overlap & GPT-5.3 H/C & 41 & 0.645 & 0.654 & +0.010 & 0.617 \\
Additional clinical concept overlap & GPT-5.4 C/A & 49 & 0.586 & 0.619 & +0.033 & 0.013 \\
\midrule
Named levels-of-care overlap & GPT-5.3 C/A & 47 & 0.664 & 0.696 & +0.032 & 0.342 \\
Named levels-of-care overlap & GPT-5.3 H/A & 40 & 0.710 & 0.727 & +0.017 & 0.734 \\
Named levels-of-care overlap & GPT-5.3 H/C & 40 & 0.681 & 0.723 & +0.042 & 0.067 \\
Named levels-of-care overlap & GPT-5.4 C/A & 48 & 0.714 & 0.740 & +0.026 & 0.313 \\
\bottomrule
\end{tabular*}
\end{table}




\end{appendices}


\end{document}